\def\year{2020}\relax
\documentclass[letterpaper]{article} % DO NOT CHANGE THIS
\usepackage{aaai20}  % DO NOT CHANGE THIS
\usepackage{times}  % DO NOT CHANGE THIS
\usepackage{helvet} % DO NOT CHANGE THIS
\usepackage{courier}  % DO NOT CHANGE THIS
\usepackage[hyphens]{url}  % DO NOT CHANGE THIS
\usepackage{graphicx} % DO NOT CHANGE THIS
\usepackage{tabularx}
\usepackage{graphicx}  % DO NOT CHANGE THIS
\usepackage{multirow}
\usepackage{enumitem}
\usepackage{tabularx}
\usepackage{booktabs} % To thicken table lines

\usepackage{amsmath}
\usepackage{amssymb}
\usepackage{subfig}
\usepackage{xcolor}
\newcommand{\citet}[1]{\citeauthor{#1}~\shortcite{#1}}
\title{Schema-Guided Dialogue State Tracking Task at DSTC8}
\author{Abhinav Rastogi, Xiaoxue Zang, Srinivas Sunkara, Raghav Gupta, Pranav Khaitan\\Google Research, Mountain View, California, USA\\
\{abhirast, xiaoxuez, srinivasksun, raghavgupta, pranavkhaitan\}@google.com % email address must be in roman text type, not monospace or sans serif
}
 
\begin{document}

\maketitle

\begin{abstract}
This paper gives an overview of the \textit{Schema-Guided Dialogue State Tracking} task of the 8th Dialogue System Technology Challenge. The goal of this task is to develop dialogue state tracking models suitable for large-scale virtual assistants, with a focus on data-efficient joint modeling across domains and zero-shot generalization to new APIs. This task provided a new dataset consisting of over 16000 dialogues in the training set spanning 16 domains to highlight these challenges, and a baseline model capable of zero-shot generalization to new APIs. Twenty-five teams participated, developing a range of neural network models, exceeding the performance of the baseline model by a very high margin. The submissions incorporated a variety of pre-trained encoders and data augmentation techniques. This paper describes the task definition, dataset and evaluation methodology. We also summarize the approach and results of the submitted systems to highlight the overall trends in the state-of-the-art.
\end{abstract}

\section{Introduction}
Virtual assistants help users accomplish tasks including but not limited to finding flights, booking restaurants, by providing a natural language interface to services and APIs on the web. Large-scale assistants like Google Assistant, Amazon Alexa, Apple Siri, Microsoft Cortana etc. need to support a large and constantly increasing number of services, over a wide variety of domains. Consequently, recent work has focused on scalable dialogue systems that can handle tasks across multiple application domains. Data-driven deep learning based approaches for multi-domain modeling have shown promise, both for end-to-end and modular systems involving dialogue state tracking and policy learning. This line of work has been facilitated by the release of multi-domain dialogue corpora such as MultiWOZ \cite{budzianowski2018multiwoz}, Taskmaster-1 \cite{byrne2019taskmaster}, M2M \cite{shah2018building} and FRAMES \cite{el2017frames}.

However, building large-scale assistants, as opposed to dialogue systems managing a few APIs, poses a new set of challenges. Apart from the handling a very large variety of domains, such systems need to support heterogeneous services or APIs with possibly overlapping functionality.  It should also offer an efficient way of supporting new APIs or services, while requiring little or no additional training data. Furthermore, to reduce maintenance workload and accommodate future growth, such assistants need to be robust to changes in the API's interface or addition of new slot values. Such changes shouldn't require collection of additional training data or retraining the model.

The Schema-Guided Dialogue State Tracking task at the Eighth Dialogue System Technology Challenge explores the aforementioned challenges in context of dialogue state tracking. In a task-oriented dialogue, the dialogue state is a summary of the entire conversation till the current turn. The dialogue state is used to invoke APIs with appropriate parameters as specified by the user over the dialogue history. It is also used by the assistant to generate the next actions to continue the dialogue. DST, therefore, is a core component of virtual assistants. 

In this task, participants are required to develop innovative approaches to multi-domain dialogue state tracking, with a focus on data-efficient joint modeling across APIs and zero-shot generalization to new APIs. The task is based on the Schema-Guided Dialogue (SGD) dataset\footnote{Available at github.com/google-research-datasets/dstc8-schema-guided-dialogue}, which, to the best of our knowledge, is the largest publicly available corpus of annotated task-oriented dialogues. With over 16000 dialogues in the training set spanning 26 APIs over 16 domains, it exceeds the existing dialogue corpora in scale. SGD is the first dataset to allow multiple APIs with overlapping functionality within each domain. To adequately test generalization in zero-shot settings, the evaluation sets contain unseen services and domains. The dataset is designed to serve as an effective testbed for intent prediction, slot filling, state tracking and language generation, among other tasks in large-scale virtual assistants.

\section{Related Work}
Dialogue systems have constituted an active area of research for the past few decades. The advent of commercial personal assistants has provided further impetus to dialogue systems research. As virtual assistants incorporate diverse domains, zero-shot modeling \cite{bapna2017towards,xia2018zero,shah-etal-2019-robust}, domain adaptation and transfer learning techniques \cite{yang2017transfer,rastogi2017scalable,zhu2018concept} have been explored to support new domains in a data efficient manner.

Deep learning based approaches to DST have recently gained popularity. Some of these approaches estimate the dialogue state as a distribution over all possible slot-values \cite{henderson2014,wen2017network}  or individually score all slot-value combinations \cite{mrkvsic2017neural,zhong-etal-2018-global}. Such approaches are, however, hard to scale to real-world virtual assistants, where the set of possible values for certain slots may be very large (date, time or restaurant name) and even dynamic (movie or event name). Other approaches utilizing a dynamic vocabulary of slot values \cite{rastogi2018multi,goel2019hyst} still do not allow zero-shot generalization to new services and APIs \cite{wu-etal-2019-transferable}, since they use schema elements i.e. intents and slots as fixed class labels.

Although such systems are capable of parsing the dialogue semantics in terms of these fixed intent labels, they lack understanding of the semantics of these labels. For instance, for the user utterance ``I want to buy tickets for a movie.", such models can predict \textit{BuyMovieTickets} as the correct intent based on patterns observed in the training data, but don't model either its association with the real world action of buying movie tickets, or its similarity to the action of buying concert or theatre tickets. Furthermore, because of their dependence on a fixed schema, such models are not robust to changes in the schema, and need to be retrained as new slots or intents are added. Use of domain-specific parameters renders some approaches unsuitable for zero-shot application.
% This poses a fundamental limitation on these models, since they would still work if \textit{BuyMovieTickets} was renamed to something completely unrelated. 

\section{Task}

The primary task of this challenge is to develop multi-domain models for DST suitable for the scale and complexity of large scale virtual assistants. Supporting a wide variety of APIs or services with possibly overlapping functionality is an important requirement of such assistants. A common approach to do this involves defining a large master schema that lists all intents and slots supported by the assistant. Each service either adopts this master schema for the representation of the underlying data, or provides logic to translate between its own schema and the master schema. 

The first approach involving adoption of the master schema is not ideal if a service wishes to integrate with multiple assistants, since each of the assistants could have their own master schema. The second approach involves definition of logic for translation between master schema and the service's schema, which increases the maintenance workload. Furthermore, it is difficult to develop a master schema catering to all possible use cases.

\begin{figure}[ht]
\centering
\includegraphics[width=0.36\textwidth]{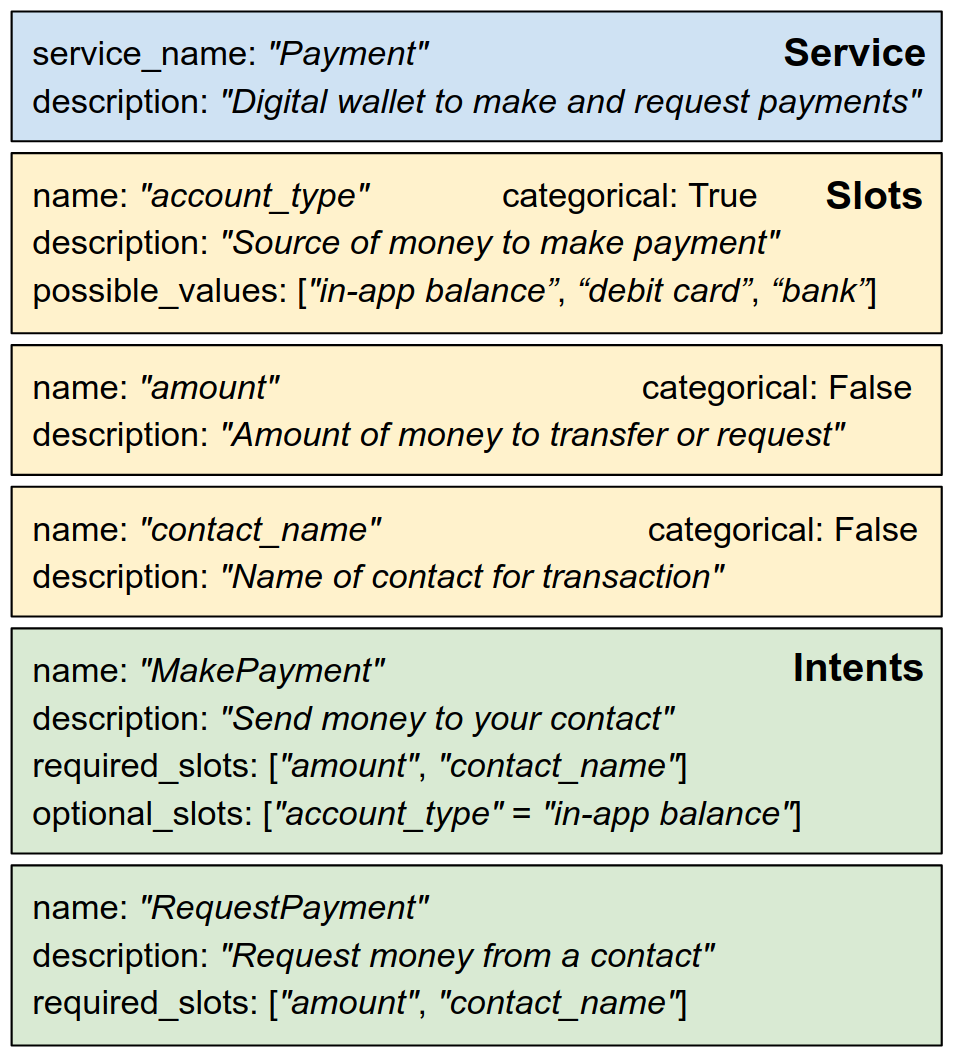}
\caption{Example schema for a digital wallet service.}
\label{fig:schema-example}
\end{figure}

\begin{figure*}[ht]
    \centering
    \includegraphics[width=0.99\textwidth]{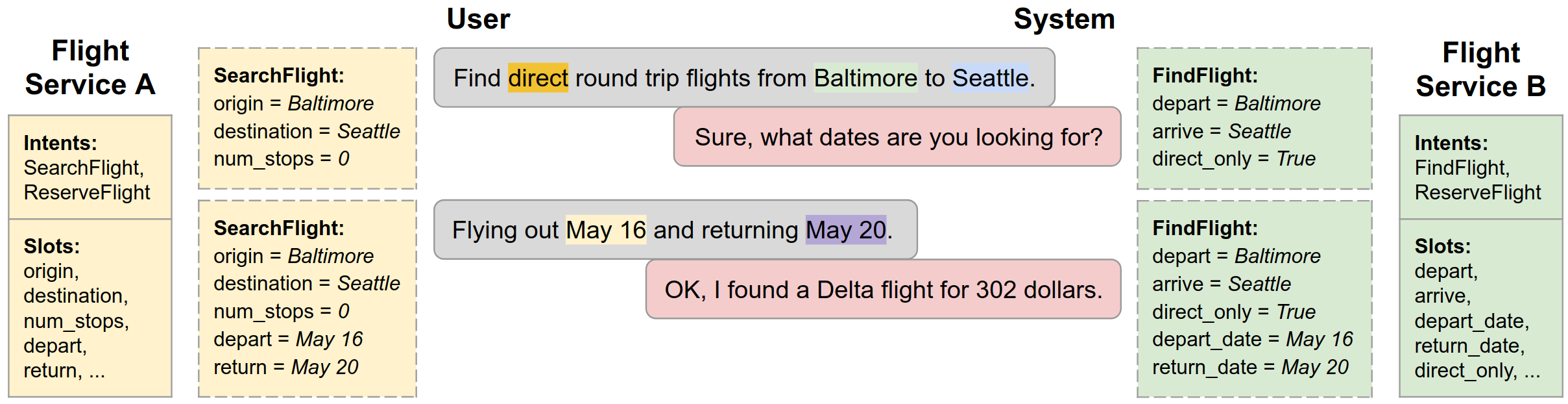}
    \caption{Dialogue state tracking labels after each user utterance in a  dialogue in the context of two different flight services. Under the schema-guided approach, the annotations are conditioned on the schema (extreme left/right) of the underlying service.}
    \label{fig:track4-model}
\end{figure*}

Additionally, while there are many similar concepts across services that can be jointly modeled, for example, the similarities in logic for querying or specifying the number of movie tickets, flight tickets or concert tickets, the master schema approach does not facilitate joint modeling of such concepts, unless an explicit mapping between them is manually defined. To address these limitations, we propose a schema-guided approach, which eliminates the need for a master schema.

\subsection{Schema-Guided Approach}

Under the Schema-Guided approach, each service provides a schema listing the supported slots and intents along with their natural language descriptions (Figure \ref{fig:schema-example} shows an example). The dialogue annotations are guided by the schema of the underlying service or API, as shown in Figure \ref{fig:track4-model}. In this example, the departure and arrival cities are captured by analogously functioning but differently named slots in both schemas. Furthermore, values for the \textit{number\_stops} and \textit{direct\_only} slots highlight idiosyncrasies between services interpreting the same concept.

The natural language descriptions present in the schema are used to obtain a semantic representation of intents and slots. The assistant employs a single unified model containing no domain or service specific parameters to make predictions conditioned on these schema elements. Using a single model facilitates representation and transfer of common knowledge across related concepts in different services. Since the model utilizes semantic representation of schema elements as input, it can interface with unseen services or APIs on which it has not been trained. It is also robust to changes like the addition of new intents or slots to the service. In addition, the participants are allowed to use any external datasets or resources to bootstrap their models.

\section{Dataset}

As shown in Table \ref{table:datasets}, our Schema-Guided Dialogue (SGD) dataset exceeds other datasets in most of the metrics at scale. The especially larger number of domains, slots, and slot values, and the presence of multiple services per domain, are representative of these scale-related challenges. Furthermore, our evaluation sets contain many services, and consequently slots, which are not present in the training set, to help evaluate model performance on unseen services.

\subsection{Data Representation}

The dataset consists of conversations between a virtual assistant and a user. Each conversation can span multiple services across various domains. The dialogue is represented as a sequence of turns, each containing a user or system utterance. The annotations for each turn are grouped into frames, where each frame corresponds to a single service. The annotations for user turns include the active intent, the dialogue state and slot spans for the different slots values mentioned in the turn. For system turns, we have the system actions representing the semantics of the system utterance. Each system action is represented using a dialogue act with optional parameters.

In addition to the dialogues, for each service used in the dataset, a normalized representation of the interface exposed is provided as the schema. The schema contains details like the name of the service, the list of tasks supported by the service (intents) and the attributes of the entities used by the service (slots). The schema also contains natural language descriptions of the service, intents and slots which can be used for developing models which can condition their predictions on the schema.

\subsection{Comparison With Other Datasets}

To reflect the constraints present in real-world services and APIs, we impose a few constraints on the data. Our dataset does not expose the set of all possible values for certain slots. Having such a list is impractical for slots like date or time because they have infinitely many possible values or for slots like movie or song names, for which new values are periodically added. Such slots are specifically identified as non-categorical slots. In our evaluation sets, we ensured the presence of a significant number of values which were not previously seen in the training set to evaluate the performance of models on unseen values. Some slots like gender, number of people, etc. are classified as categorical and we provide a list of all possible values for them. However, these values are assumed to be not consistent across services. E.g., different services may use (`male', `female'), (`M', `F') or (`he', `she') as possible values for gender slot.

Real-world services can only be invoked with certain slot combinations: e.g. most restaurant reservation APIs do not let the user search for restaurants by date without specifying a location. Although this constraint has no implications on the dialogue state tracking task, it restricts the possible conversational flows. Hence, to prevent flows not supported by actual services, we restrict services to be called with a list of slot combinations. The different service calls supported by a service are listed as intents with each intent specifying a list of required slots. The intent cannot be called without providing values for these required slots. Each intent also contains a list of optional slots with default values which can be overridden by the user.

In our dataset, we also have multiple services per domain with overlapping functionality. The intents across these services are similar but differ in terms of intent names, intent arguments, slot names, etc. In some cases, there is no one to one mapping between slot names (e.g., the \textit{num\_stops} and \textit{direct\_only} slots in Figure \ref{fig:track4-model}). With an ever increasing number of services and service providers, we believe that having multiple similar services per domain is much closer to the situation faced by virtual assistants than having one unique service per domain. 

% Current dialogue systems need a considerable amount of training data to achieve a good performance, which makes it hard to bootstrap a dialogue system for a new domain with limited data. Recent work has focused on zero-shot modeling \cite{bapna2017towards,xia2018zero,shah-etal-2019-robust}, domain adaptation and transfer learning approaches \cite{rastogi2017}. In our dataset we have 3 domains and 14 services in the test set which are not present in the training dataset. This forces the models to generalize to domains and services not seen before. We also provide natural language descriptions of the different services, intents and slots in the schema to enable models to easily condition on the pertinent domain and also enable the model to recognize semantics of intents and identify patterns across domains.

\subsection{Data Collection And Dataset Analysis}\label{sec:data-collect}

Our data collection setup uses a dialogue simulator to generate dialogue outlines first and then paraphrase them to obtain natural utterances. Using a dialogue simulator offers us multiple advantages. First, it ensures the coverage of a large variety of dialogue flows by filtering out similar flows in the simulation phase, thus creating a much diverse dataset. 
%To ensure naturalness of the generated conversations, we used the conversational flows present in other public datasets like MultiWOZ 2.0 \cite{budzianowski2018multiwoz} and WOZ2.0 \cite{wen2017network} as a guideline while developing the dialogue simulator. 
Second, simulated dialogues do not require manual annotation, as opposed to a Wizard-of-Oz setup \cite{kelley1984iterative}, which is a common approach utilized in other datasets~\cite{budzianowski2018multiwoz}. It has been shown that such datasets suffer from substantial annotation errors~\cite{eric2019multiwoz}. Thirdly, using a simulator greatly simplifies the data collection task and instructions as only paraphrasing is needed to achieve a natural dialogue. This is particularly important for creating a large dataset spanning multiple domains.

The 20 domains present across the train, dev and test datasets are listed in Table \ref{table:domains}, as are the details regarding which domains are present in each of the datasets. We create synthetic implementations of a total of 45 services or APIs over these domains. Our simulator framework interacts with these services to generate dialogue outlines, which are structured representations of dialogue semantics. We then use a crowd-sourcing procedure to paraphrase these outlines to natural language utterances. Our novel crowd-sourcing procedure preserves all annotations obtained from the simulator and does not require any extra annotations after dialogue collection. In this section, we describe these steps briefly and then present analyses of the collected dataset. 

All the services are implemented using a SQL engine. Since entity attributes are often correlated, we decided not to sample synthetic entities and instead relied on sampling entities from Freebase. The dialogue simulator interacts with the services to generate valid dialogue outlines. The simulator consists of two agents  playing  the  roles  of  the  user  and  the  system.  Both agents interact with each other using a finite set of actions specified through dialogue acts over a probabilistic automaton designed to capture varied dialogue trajectories. At the start of the conversation, the user agent is seeded with a scenario, which is a sequence of intents to be fulfilled. The user agent generates dialogue acts to be output and combines them with values retrieved from the service/API to create the user actions. The system agent responds by following a similar procedure but also ensures that the generated flows are valid.  We identified over 200 distinct scenarios for the training set each consisting up to 5 intents from various domains. Finally, the dialogue outlines generated are paraphrased into a natural conversation by crowd workers. We ensure that the annotations for the dialogue state and slots generated by the simulator are preserved and hence need no other annotation. We omit details for brevity: please refer to \citet{rastogi2019towards} for more details.

\begin{figure}
    \centering
    \subfloat[Histogram of lengths of training set dialogues.]{  \includegraphics[width=0.95\columnwidth]{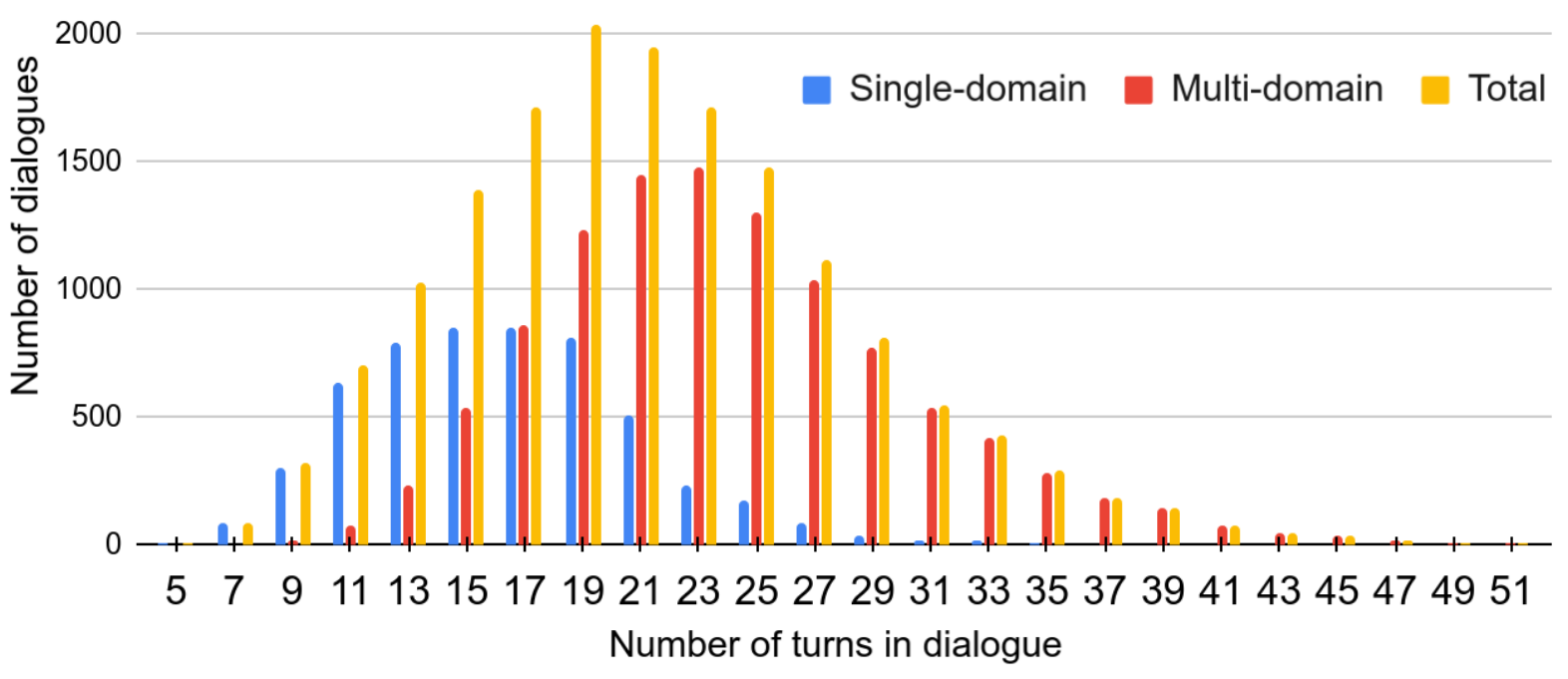} \label{fig:dialogue_lengths}} \qquad
    \subfloat[Distribution of dialogue acts in training set.]{\includegraphics[width=0.95\columnwidth]{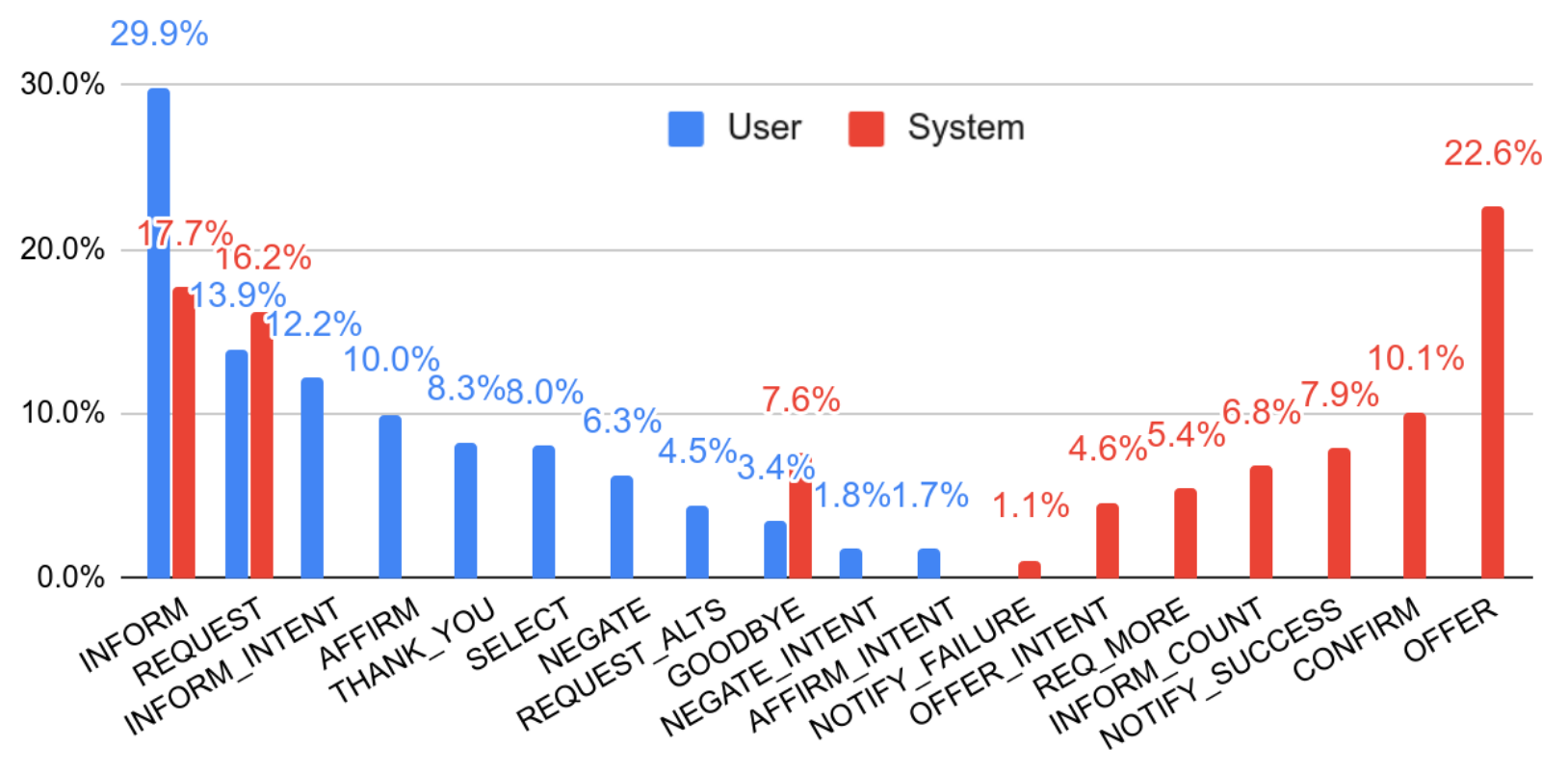}
    \label{fig:dialogue_act_distribution}}
\caption{Detailed statistics of the SGD dataset.}
\end{figure}

The entire dataset consists of over 16K dialogues spanning multiple domains. Overall statistics of the dataset and comparison with other datasets can be seen in Table \ref{table:datasets}. Figure \ref{fig:dialogue_lengths} shows the details of the distribution of dialogue lengths across single-domain and multi-domain dialogues. The single-domain dialogues in our dataset contain an average of 15.3 turns, whereas the multi-domain ones contain 23 turns on average. Figure \ref{fig:dialogue_act_distribution} shows the frequency of the different dialogue acts contained in the dataset. The dataset also contains a significant number of unseen domains/APIs in the dev and test sets. 77\% of the dialogue turns in the test set and 45\% of the turns in dev set contain at least one service not present in the training set. This facilitates the development of models which can generalize to new domains with very few labelled examples.

\begin{table*}[t!]
\centering
    \begin{tabular}[t]{ c | c c c c c c }

    \textbf{Metric $\downarrow$ Dataset $\rightarrow$} & \textbf{DSTC2} & \textbf{WOZ2.0} & \textbf{FRAMES} & \textbf{M2M} & \textbf{MultiWOZ} & \textbf{SGD} \\\hline
    No. of domains & 1 & 1 & 3 & 2 & 7 & \textbf{16}\\
    No. of dialogues &  1,612 & 600 & 1,369 & 1,500 & 8,438 & \textbf{16,142}\\
    Total no. of turns &  23,354 & 4,472 & 19,986 & 14,796 & 113,556 & \textbf{329,964}\\
    Avg. turns per dialogue & 14.49 & 7.45 & 14.60 & 9.86 & 13.46 & \textbf{20.44} \\
    Avg. tokens per turn & 8.54 & 11.24 & 12.60 & 8.24 & \textbf{13.13} & 9.75\\
    Total unique tokens & 986 & 2,142 & 12,043 & 1,008 & 23,689 & \textbf{30,352}\\
    No. of slots &  8 & 4 & 61 & 13 & 24 & \textbf{214}\\
    No. of slot values & 212 & 99 & 3,871 & 138 & 4,510 & \textbf{14,139}\\

    \end{tabular}
    \caption{Comparison of our SGD dataset to existing related datasets for task-oriented dialogue. Note that the numbers reported are for the training portions for all datasets except FRAMES, where the numbers for the complete dataset are reported.}
    \label{table:datasets}
\end{table*}

\begin{table}[!htb]
    \centering
    \def\arraystretch{1.23}
    \resizebox{\columnwidth}{!}{
        \begin{tabular}{ l | cc||l | cc } 
            \textbf{Domain} & \textbf{\#Intents}  & \textbf{\#Dialogs} & \textbf{Domain} & \textbf{\#Intents}  & \textbf{\#Dialogs} \\ \hline
            Alarm$^{2,3}$ & 2 (1) & 324 & Movies$^{1,2,3}$ & 5 (3) & 2339\\
            Banks$^{1,2}$ & 4 (2) & 1021 & Music$^{1,2,3}$ & 6 (3) & 1833\\
            Buses$^{1,2,3}$ & 6 (3) & 3135 & Payment$^3$ & 2 (1) & 222\\
            Calendar$^1$ & 3 (1) & 1602 & RentalCars$^{1,2,3}$ & 6 (3) & 2510\\
            Events$^{1,2,3}$ & 7 (3) & 4519 & Restaurants$^{1,2,3}$ & 4 (2) & 3218\\
            Flights$^{1,2,3}$ & 10 (4) & 3644 & RideSharing$^{1,2,3}$ & 2 (2) & 2223\\
            Homes$^{1,2,3}$ & 2 (1) & 1273  & Services$^{1,2,3}$ & 8 (4) & 2956\\
            Hotels$^{1,2,3}$ & 8 (4) & 4992  & Train$^{3}$ & 2 (1) & 350\\
            Media$^{1,2,3}$ & 6 (3) & 1656 & Travel$^{1,2,3}$ & 1 (1) & 2808\\
            Messaging$^3$ & 1 (1) & 298 & Weather$^{1,2,3}$ & 1 (1) & 1783\\
          \end{tabular}
    }
    \caption{The total number of intents (services in parentheses) and dialogues for each domain across train$^1$, dev$^2$ and test$^3$ sets. Superscript indicates the datasets in which dialogues from the domain are present. Multi-domain dialogues contribute to counts of each domain. The domain Services includes salons, dentists, doctors, etc.}
    \label{table:domains}
\end{table}

\section{Submissions}

\begin{table*}[ht] 
\centering
    \setlength\tabcolsep{2pt}
    \def\arraystretch{1.35}
\resizebox{2\columnwidth}{!}{
\begin{tabular}{c!{\vrule width 2pt}c|c|c|c!{\vrule width 2pt}c|c|c|c!{\vrule width 2pt}c|c|c|c}

\multirow{2}{*}{\textbf{Team name}} & 
  \multicolumn{4}{c!{\vrule width 2pt}}{\textbf{All services}} &
  \multicolumn{4}{c!{\vrule width 2pt}}{\textbf{Seen services}} &
  \multicolumn{4}{c}{\textbf{Unseen services}} \\ \cline{2-13}
& \textbf{Joint GA} & \textbf{Avg GA} & \textbf{Intent Acc} & \textbf{Req Slot F1} & \textbf{Joint GA} & \textbf{Avg GA} & \textbf{Intent Acc} & \textbf{Req Slot F1} &  \textbf{Joint GA} & \textbf{Avg GA} & \textbf{Intent Acc} &  \textbf{Req Slot F1} \\ \hline 
\textbf{Team 9*} & \textbf{0.8653} & \textbf{0.9697} & 0.9482 & 0.9847 & \textbf{0.9241} & \textbf{0.9799} & 0.9571 & 0.9936 & \textbf{0.8456} & \textbf{0.9662} & 0.9452 & 0.9817 \\\hline
Team 14 & 0.7726 & 0.9217 & \textbf{0.9674} & 0.9932 & 0.9005 & 0.9606 & 0.9578 & 0.9963 & 0.7299 & 0.9081 & \textbf{0.9706} & 0.9921 \\\hline
Team 12* & 0.7375 & 0.9199 & 0.9234 & \textbf{0.9948} & 0.8795 & 0.9566 & 0.9581 & 0.9965 & 0.6901 & 0.9071 & 0.9118 & \textbf{0.9943} \\\hline
Team 8 & 0.7344 & 0.9251 & N.A. & 0.8713 & 0.9106 & 0.9708 & N.A. & 0.8475 & 0.6757 & 0.9093 & N.A. & 0.8793 \\\hline
Team 5* & 0.7303 & 0.9249 & 0.9426 & 0.9814 & 0.8936 & 0.9662 & \textbf{0.9594} & 0.9920 & 0.6758 & 0.9105 & 0.9370 & 0.9779 \\\hline
Team 10 & 0.6946 & 0.9105 & 0.9509 & 0.8713 & 0.9203 & 0.9780 & 0.9560 & 0.8475 & 0.6193 & 0.8871 & 0.9492 & 0.8793 \\\hline
Team 13 & 0.6616 & 0.9037 & 0.9368 & 0.9854 & 0.8584 & 0.9527 & 0.9534 & 0.9960 & 0.5960 & 0.8867 & 0.9312 & 0.9819 \\\hline
Team 7 & 0.6316 & 0.8595 & 0.9231 & 0.9797 & 0.8410 & 0.9356 & 0.9449 & 0.9951 & 0.5617 & 0.8331 & 0.9158 & 0.9746 \\\hline
Team 6 & 0.6102 & 0.8430 & 0.9041 & 0.8713 & 0.6764 & 0.8397 & 0.9483 & 0.8475 & 0.5881 & 0.8442 & 0.8893 & 0.8793 \\\hline
Team 18 & 0.6099 & 0.9049 & 0.9423 & 0.9723 & 0.8223 & 0.9601 & 0.9540 & 0.9876 & 0.5390 & 0.8858 & 0.9384 & 0.9672 \\\hline
Team 21 & 0.5475 & 0.8670 & 0.9344 & 0.8713 & 0.7514 & 0.9190 & 0.9418 & 0.8475 & 0.4795 & 0.8489 & 0.9319 & 0.8793 \\\hline
Team 16* & 0.5410 & 0.8027 & 0.9137 & 0.8713 & 0.5289 & 0.7515 & 0.9561 & 0.8475 & 0.5450 & 0.8205 & 0.8995 & 0.8793 \\\hline
Team 3 & 0.5035 & 0.7853 & 0.8789 & 0.9581 & 0.6172 & 0.8174 & 0.9565 & 0.9902 & 0.4656 & 0.7741 & 0.8530 & 0.9474 \\\hline
Team 25 & 0.4801 & 0.7706 & 0.8765 & 0.9862 & 0.5412 & 0.7659 & 0.9379 & 0.9960 & 0.4597 & 0.7722 & 0.8560 & 0.9829 \\\hline
Team 20 & 0.4774 & 0.7148 & 0.8400 & 0.9453 & 0.7847 & 0.9209 & 0.9416 & 0.9840 & 0.3748 & 0.6432 & 0.8061 & 0.9324 \\\hline
Team 23* & 0.4647 & 0.7500 & 0.7474 & 0.9703 & 0.5275 & 0.7391 & 0.8710 & 0.9710 & 0.4438 & 0.7538 & 0.7061 & 0.9700 \\\hline
Team 11 & 0.4212 & 0.7056 & 0.9070 & 0.9663 & 0.6375 & 0.8226 & 0.9397 & 0.9964 & 0.3490 & 0.6649 & 0.8961 & 0.9563 \\\hline
Team 15 & 0.3907 & 0.6874 & 0.9379 & 0.9799 & 0.4965 & 0.7357 & 0.9516 & \textbf{0.9970} & 0.3554 & 0.6706 & 0.9333 & 0.9742 \\\hline
Team 2* & 0.3647 & 0.7438 & 0.9243 & 0.9764 & 0.7363 & 0.9132 & 0.9492 & 0.9925 & 0.2406 & 0.6850 & 0.9160 & 0.9710 \\\hline
Team 22 & 0.3259 & 0.6714 & 0.9077 & 0.9525 & 0.6772 & 0.8966 & 0.7855 & 0.9504 & 0.2285 & 0.6082 & 0.9416 & 0.9530 \\\hline
Team 24 & 0.3198 & 0.6347 & 0.8764 & 0.9729 & 0.7077 & 0.8888 & 0.9413 & 0.9846 & 0.1903 & 0.5464 & 0.8548 & 0.9690 \\\hline
Team 19 & 0.3052 & 0.6302 & 0.9240 & 0.9668 & 0.5140 & 0.7476 & 0.9607 & 0.9953 & 0.2355 & 0.5894 & 0.9118 & 0.9572 \\\hline
Team 17 & 0.2525 & 0.5721 & 0.8875 & 0.9680 & 0.4179 & 0.6858 & 0.9433 & 0.9952 & 0.1973 & 0.5326 & 0.8689 & 0.9590 \\\hline
Team 1 & 0.2511 & 0.5609 & 0.8406 & 0.9648 & 0.4255 & 0.6825 & 0.9164 & 0.9949 & 0.1929 & 0.5187 & 0.8153 & 0.9547 \\\hline
Team 4 & 0.2354 & 0.5365 & 0.8841 & 0.9445 & 0.4004 & 0.6333 & 0.9228 & 0.9523 & 0.1803 & 0.5029 & 0.8712 & 0.9419 \\\hline
Baseline & 0.2537 & 0.5605 & 0.9064 & 0.9651 & 0.4125 & 0.6778 & 0.9506 & 0.9955 & 0.2000 & 0.5192 & 0.8915 & 0.9547  \\ \hline
\toprule
\end{tabular}
}
\caption{The best submission from each team, ordered by the joint goal accuracy on the test set. Teams marked with * submitted their papers to the workshop. We could not identify the teams for three of the submitted papers.}
\label{table:metrics}
\end{table*}

The submissions from 25 teams included a variety of approaches and innovative solutions to specific problems posed by this dataset. For the workshop, we received submissions from 9 of these teams. In this section, we provide a short summary of the approaches followed by these teams. For effective generalization to unseen APIs, most teams used pre-trained encoders to encode schema element descriptions. Unless otherwise mentioned, a pre-trained BERT \cite{devlin2019bert} encoder was used.

\begin{itemize}
    \item \textbf{Team 2 \cite{lo2020dstc}:} This was the only paper not using a pre-trained encoder, thus providing another important baseline. They rely on separate RNNs to encode service, slot and intent descriptions, and a BiRNN to encode dialogue history. Slot values are inferred using a TRADE-like encoder-decoder setup with a 3-way slot status gate, using the utterance encoding and schema element embeddings as context.
    
    \item \textbf{Team 5 \cite{lei2020dstc}:} They predict values for categorical slots using a softmax over all candidate values. Non-categorical slot values are predicted by first predicting the status of each slot and then using a BiLSTM-CRF layer for BIO tagging \cite{ramshaw1995text}. They also utilize a slot adoption tracker to predict if the values proposed by the system are accepted by the user.

    \item \textbf{Team 9 \cite{ma2020dstc}:} This team submitted the winning entry, beating the second-placed team by around 9\% in terms of joint goal accuracy. They use two separate models for categorical and non-categorical slots, and treat numerical categorical slots as non-categorical. They also use the entire dialogue history as input. They perform data augmentation by back translation between English and Chinese, which seems to be one of the distinguishing factors resulting in a much higher accuracy.
    
    \item \textbf{Team 12 \cite{ruan2020dstc}:} They use auxiliary binary features to connect previous intent to current intent, slots to dialogue history and source slots to target slots for slot transfer. Non-categorical slots are modeled similar to question answering by adding a null token and predicting spans for slot values. In-domain and cross-domain slot transfers are modeled as separate binary decisions by passing the slot descriptions as additional inputs.

    \item \textbf{Team 16 \cite{shi2020dstc}:} They convert the tracking task for both categorical and non-categorical slots into a question answering task by feeding in the schema and the previous turns as the context. Similar to the baseline model, prediction is performed in two stages. The status of each slot (active/inactive/dontcare) is predicted using a classifier, following which the value is predicted as a span in the context. The same network is used for the different prediction tasks but the leading token and separator tokens used are different. They observe large gains by fine-tuning the schema embeddings and increasing the number of past turns fed as context.
    
    \item \textbf{Team 23 \cite{gulyaev2020dstc}:} They use a large scale multi-task model utilizing a single pass of a BERT based model for all tasks. Embeddings are calculated for the intents and slot value by using dialogue history, service and slot descriptions, possible values for categorical slots and are used for the various predictions.
    
    \item \textbf{Anonymous Team A \cite{balaraman2020dstc}:} We could not identify which team submitted this model. They use multi-head attention twice to obtain domain-conditioned and slot-conditioned representations of the dialogue history. These representations are concatenated to obtain the full context which is used for the various predictions.

    \item \textbf{Anonymous Team B \cite{li2020dstc}:} We could not identify which team submitted this model. They use separate NLU systems for the sub tasks of predicting intents, requested slots, slot status, categorical and non-categorical slot values. They use a rule-based DST system with a few additions resulting in significant improvement. The improvements include adding dropout to intent prediction to account for train-test mismatch, using the entire predicted slot status distribution and separate binary predictions for slot transfer.
    
    \item \textbf{Anonymous Team C \cite{zheng2020dstc}:} They use a two-stage model with a candidate tracker for NLU and a candidate classifier to update the dialogue state. A slot tagger identifies slot values, which are used to update the candidate tracker. The candidate classifier uses the utterances and slot/intent descriptions to predict the final dialogue state. They also use an additional loss to penalize incorrect prediction on which slots appear in the current turn.

\end{itemize}

\section{Evaluation}
We consider the following metrics for automatic evaluation of different submissions. Joint goal accuracy has been used as the primary metric to rank the submissions.

\begin{enumerate}
    \item \textbf{Active Intent Accuracy:} The fraction of user turns for which the active intent has been correctly predicted.
    \item \textbf{Requested Slot F1:} The macro-averaged F1 score for requested slots over all eligible turns. Turns with no requested slots in ground truth and predictions are skipped.
    \item \textbf{Average Goal Accuracy:} For each turn, we predict a single value for each slot present in the dialogue state. This is the average accuracy of predicting the value of a slot correctly.
    \item \textbf{Joint Goal Accuracy:} This is the average accuracy of predicting \textit{all} slot assignments for a given service in a turn correctly. 
\end{enumerate}

In order to better reflect model performance in our task's specific setting, we introduce changes in the definitions of evaluation metrics from prior work. These are listed below:
\begin{itemize}[leftmargin=*]
    \item \textbf{Joint goal accuracy calculation:} Traditionally, joint goal accuracy has been defined as the accuracy of predicting the dialogue state for all domains correctly. This is not practical in our setup, as the large number of services would result in near zero joint goal accuracy if the traditional definition is used. Furthermore, an incorrect dialogue state prediction for a service in the beginning of a dialogue degrades the joint goal accuracy for all future turns, even if the predictions for all other services are correct. Hence, joint goal accuracy calculated this way may not provide as much insight into the performance on different services. To address these concerns, only the services which are active or pertinent in a turn are included in the dialogue state. Thus, a service ceases to be a part of the dialogue state once its intent has been fulfilled. 

    \item \textbf{Fuzzy matching for non-categorical slot values:} The presence of non-categorical slots is another distinguishing feature of our dataset. These slots don't have a predefined vocabulary, and their values are predicted as a substring or span of the past user or system utterances. Drawing inspiration from the metrics used for slot tagging in spoken language understanding, we use a fuzzy matching score for non-categorical slots to reward partial matches with the ground truth.

    \item \textbf{Average goal accuracy:} To calculate average goal accuracy, we do not take into account instances when both the ground truth and the predicted values for a slot are empty. Since for a given slot, a large number of utterances have an empty assignment, models can achieve a relatively high average goal accuracy just by predicting an empty assignment for each slot unless specifically excluded as in our evaluation.
\end{itemize}

\section{Results}

The test set contains a total of 21 services, among which 6 services are also present in the training set (seen services), whereas the remaining 15 are not present in the training set (unseen services). Table \ref{table:metrics} shows the evaluation metrics for the different submissions obtained on the test set. It also lists the performance of different submissions on seen and unseen services, helping evaluate the effectiveness in zero-shot settings. Team 9 achieved a very high joint goal accuracy of \textbf{86.53\%}, around 9\% higher than the second-placed team. We observed the following trends across submissions:

\begin{itemize}
    \item For unseen services, performance on categorical slots is comparable to that on non-categorical slots. On the other hand, for seen services, the performance on categorical slots is better. This could be because there is less signal to differentiate between the different possible values for a categorical slot when they have not been observed in the training set.
    
    \item The winning team's performance on seen services is similar to that of the other top teams. However, the winning team has a considerable edge on unseen services, outperforming the second team by around 12\% in terms of joint goal accuracy. This margin was observed across both categorical and non-categorical slots.
    
    \item Among unseen services, when looking at services belonging to unseen domains, the winning team was ahead of the other teams by at least 15\%. The performance on categorical slots for unseen domains was about the same as that for seen services and domains. For other teams, there was at least a 20\% drop in accuracy of categorical slots in unseen domains vs seen domains and services.
    
    \item The joint goal accuracy of most of the models was worse by 15 percentage points on an average on the test set as compared to the dev set. This could be because the test set contains a much higher proportion of turns with at least one unseen services as compared to the dev set (77\% and 45\% respectively).
\end{itemize}

\section{Summary}
In this paper, we summarized the Schema-Guided Dialogue State Tracking task conducted at the Eighth Dialogue System Technology Challenge. This task challenged participants to develop dialogue state tracking models for large scale virtual assistants, with particular emphasis on joint modeling across different domains and APIs for data-efficiency and zero-shot generalization to new/unseen APIs. In order to encourage the development of such models, we constructed a new dataset spanning 16 domains (and 4 new domains in dev and test sets), defining multiple APIs with overlapping functionality for each of these domains. We advocated the use of schema-guided approach to building large-scale assistants, facilitating data-efficient joint modeling across domains while reducing maintenance workload.

The Schema-Guided Dialogue dataset released as part of this task is the first to highlight many of the aforementioned challenges. As a result, this task led to the development of several models utilizing the schema-guided approach for dialogue state tracking. The models extensively utilized pre-trained encoders like BERT \cite{devlin2019bert}, XLNet \cite{yang2019xlnet} etc. and employed data augmentation techniques to achieve effective zero-shot generalization to new APIs. The proposed schema-guided approach is fairly general and can be used to develop other dialogue system components such as language understanding, policy and response generation. We plan to explore them in future works.

\subsubsection{Acknowledgements} The authors thank Guan-Lin Chao, Amir Fayazi and Maria Wang for their advice and assistance.

\fontsize{9.0pt}{10.0pt} \selectfont

\bibliography{aaai}

\begin{thebibliography}{}

\bibitem[\protect\citeauthoryear{Balaraman and
  Magnini}{2020}]{balaraman2020dstc}
Balaraman, V., and Magnini, B.
\newblock 2020.
\newblock Domain-aware dialogue state tracker for multi-domain dialogue
  systems.
\newblock In {\em Dialog System Technology Challenge Workshop at AAAI}.

\bibitem[\protect\citeauthoryear{Bapna \bgroup et al\mbox.\egroup
  }{2017}]{bapna2017towards}
Bapna, A.; T{\"{u}}r, G.; Hakkani{-}T{\"{u}}r, D.; and Heck, L.~P.
\newblock 2017.
\newblock Towards zero-shot frame semantic parsing for domain scaling.
\newblock In {\em Interspeech 2017, 18th Annual Conference of the International
  Speech Communication Association, Stockholm, Sweden, August 20-24, 2017}.

\bibitem[\protect\citeauthoryear{Budzianowski \bgroup et al\mbox.\egroup
  }{2018}]{budzianowski2018multiwoz}
Budzianowski, P.; Wen, T.-H.; Tseng, B.-H.; Casanueva, I.; Ultes, S.; Ramadan,
  O.; and Gasic, M.
\newblock 2018.
\newblock Multiwoz-a large-scale multi-domain wizard-of-oz dataset for
  task-oriented dialogue modelling.
\newblock In {\em Proceedings of the 2018 Conference on Empirical Methods in
  Natural Language Processing},  5016--5026.

\bibitem[\protect\citeauthoryear{Byrne \bgroup et al\mbox.\egroup
  }{2019}]{byrne2019taskmaster}
Byrne, B.; Krishnamoorthi, K.; Sankar, C.; Neelakantan, A.; Goodrich, B.;
  Duckworth, D.; Yavuz, S.; Dubey, A.; Kim, K.-Y.; and Cedilnik, A.
\newblock 2019.
\newblock Taskmaster-1: Toward a realistic and diverse dialog dataset.
\newblock In {\em Proceedings of the 2019 Conference on Empirical Methods in
  Natural Language Processing and the 9th International Joint Conference on
  Natural Language Processing (EMNLP-IJCNLP)},  4506--4517.

\bibitem[\protect\citeauthoryear{Devlin \bgroup et al\mbox.\egroup
  }{2019}]{devlin2019bert}
Devlin, J.; Chang, M.-W.; Lee, K.; and Toutanova, K.
\newblock 2019.
\newblock Bert: Pre-training of deep bidirectional transformers for language
  understanding.
\newblock In {\em Proceedings of the 2019 Conference of the North American
  Chapter of the Association for Computational Linguistics: Human Language
  Technologies, Volume 1 (Long and Short Papers)},  4171--4186.

\bibitem[\protect\citeauthoryear{El~Asri \bgroup et al\mbox.\egroup
  }{2017}]{el2017frames}
El~Asri, L.; Schulz, H.; Sharma, S.; Zumer, J.; Harris, J.; Fine, E.; Mehrotra,
  R.; and Suleman, K.
\newblock 2017.
\newblock Frames: a corpus for adding memory to goal-oriented dialogue systems.
\newblock In {\em Proceedings of the 18th Annual SIGdial Meeting on Discourse
  and Dialogue},  207--219.

\bibitem[\protect\citeauthoryear{Eric \bgroup et al\mbox.\egroup
  }{2019}]{eric2019multiwoz}
Eric, M.; Goel, R.; Paul, S.; Sethi, A.; Agarwal, S.; Gao, S.; and Hakkani-Tur,
  D.
\newblock 2019.
\newblock Multiwoz 2.1: Multi-domain dialogue state corrections and state
  tracking baselines.
\newblock {\em arXiv preprint arXiv:1907.01669}.

\bibitem[\protect\citeauthoryear{Goel, Paul, and
  Hakkani-T{\"u}r}{2019}]{goel2019hyst}
Goel, R.; Paul, S.; and Hakkani-T{\"u}r, D.
\newblock 2019.
\newblock Hyst: A hybrid approach for flexible and accurate dialogue state
  tracking.
\newblock {\em arXiv preprint arXiv:1907.00883}.

\bibitem[\protect\citeauthoryear{Gulyaev \bgroup et al\mbox.\egroup
  }{2020}]{gulyaev2020dstc}
Gulyaev, P.; Elistratova, E.; Konovalov, V.; Kuratov, Y.; Pugachev, L.; and
  Burtsev, M.
\newblock 2020.
\newblock Goal-oriented multi-task bert-based dialogue state tracker.
\newblock In {\em Dialog System Technology Challenge Workshop at AAAI}.

\bibitem[\protect\citeauthoryear{Henderson, Thomson, and
  Young}{2014}]{henderson2014}
Henderson, M.; Thomson, B.; and Young, S.
\newblock 2014.
\newblock Word-based dialog state tracking with recurrent neural networks.
\newblock In {\em Proceedings of the 15th Annual Meeting of the Special
  Interest Group on Discourse and Dialogue (SIGDIAL)},  292--299.

\bibitem[\protect\citeauthoryear{Kelley}{1984}]{kelley1984iterative}
Kelley, J.~F.
\newblock 1984.
\newblock An iterative design methodology for user-friendly natural language
  office information applications.
\newblock {\em ACM Transactions on Information Systems (TOIS)} 2(1):26--41.

\bibitem[\protect\citeauthoryear{Lei \bgroup et al\mbox.\egroup
  }{2020}]{lei2020dstc}
Lei, S.; Liu, S.; Sen, M.; Jiang, H.; and Wang, X.
\newblock 2020.
\newblock Zero-shot state tracking and user adoption tracking on schema-guided
  dialogue.
\newblock In {\em Dialog System Technology Challenge Workshop at AAAI}.

\bibitem[\protect\citeauthoryear{Li, Xiong, and Cao}{2020}]{li2020dstc}
Li, M.; Xiong, H.; and Cao, Y.
\newblock 2020.
\newblock The sppd system for schema guided dialogue state tracking challenge.
\newblock In {\em Dialog System Technology Challenge Workshop at AAAI}.

\bibitem[\protect\citeauthoryear{Lo \bgroup et al\mbox.\egroup
  }{2020}]{lo2020dstc}
Lo, K.-L.; Lu, T.-W.; Weng, T.-t.; Chen, I.-H.; and Chen, Y.-N.
\newblock 2020.
\newblock Lion-net: Lightweight ontology-independent networks for schema-guided
  dialogue state generation.
\newblock In {\em Dialog System Technology Challenge Workshop at AAAI}.

\bibitem[\protect\citeauthoryear{Ma \bgroup et al\mbox.\egroup
  }{2020}]{ma2020dstc}
Ma, Y.; Zeng, Z.; Zhu, D.; Li, X.; Yang, Y.; Yao, X.; Zhou, K.; and Shen, J.
\newblock 2020.
\newblock An end-to-end dialogue state tracking system with machine reading
  comprehension and wide \& deep classification.
\newblock In {\em Dialog System Technology Challenge Workshop at AAAI}.

\bibitem[\protect\citeauthoryear{Mrk{\v{s}}i{\'c} \bgroup et al\mbox.\egroup
  }{2017}]{mrkvsic2017neural}
Mrk{\v{s}}i{\'c}, N.; S{\'e}aghdha, D.~{\'O}.; Wen, T.-H.; Thomson, B.; and
  Young, S.
\newblock 2017.
\newblock Neural belief tracker: Data-driven dialogue state tracking.
\newblock In {\em Proceedings of the 55th Annual Meeting of the Association for
  Computational Linguistics (Volume 1: Long Papers)}, volume~1,  1777--1788.

\bibitem[\protect\citeauthoryear{Ramshaw and Marcus}{1995}]{ramshaw1995text}
Ramshaw, L., and Marcus, M.
\newblock 1995.
\newblock Text chunking using transformation-based learning.
\newblock In {\em Third Workshop on Very Large Corpora}.

\bibitem[\protect\citeauthoryear{Rastogi \bgroup et al\mbox.\egroup
  }{2019}]{rastogi2019towards}
Rastogi, A.; Zang, X.; Sunkara, S.; Gupta, R.; and Khaitan, P.
\newblock 2019.
\newblock Towards scalable multi-domain conversational agents: The
  schema-guided dialogue dataset.
\newblock {\em arXiv preprint arXiv:1909.05855}.

\bibitem[\protect\citeauthoryear{Rastogi, Gupta, and
  Hakkani-Tur}{2018}]{rastogi2018multi}
Rastogi, A.; Gupta, R.; and Hakkani-Tur, D.
\newblock 2018.
\newblock Multi-task learning for joint language understanding and dialogue
  state tracking.
\newblock In {\em Proceedings of the 19th Annual SIGdial Meeting on Discourse
  and Dialogue},  376--384.

\bibitem[\protect\citeauthoryear{Rastogi, Hakkani-T{\"u}r, and
  Heck}{2017}]{rastogi2017scalable}
Rastogi, A.; Hakkani-T{\"u}r, D.; and Heck, L.
\newblock 2017.
\newblock Scalable multi-domain dialogue state tracking.
\newblock In {\em 2017 IEEE Automatic Speech Recognition and Understanding
  Workshop (ASRU)},  561--568.
\newblock IEEE.

\bibitem[\protect\citeauthoryear{Ruan \bgroup et al\mbox.\egroup
  }{2020}]{ruan2020dstc}
Ruan, Y.-P.; Ling, Z.-H.; Gu, J.-C.; and Liu, Q.
\newblock 2020.
\newblock Fine-tuning bert for schema-guided zero-shot dialogue state tracking.
\newblock In {\em Dialog System Technology Challenge Workshop at AAAI}.

\bibitem[\protect\citeauthoryear{Shah \bgroup et al\mbox.\egroup
  }{2018}]{shah2018building}
Shah, P.; Hakkani-T{\"u}r, D.; T{\"u}r, G.; Rastogi, A.; Bapna, A.; Nayak, N.;
  and Heck, L.
\newblock 2018.
\newblock Building a conversational agent overnight with dialogue self-play.
\newblock {\em arXiv preprint arXiv:1801.04871}.

\bibitem[\protect\citeauthoryear{Shah \bgroup et al\mbox.\egroup
  }{2019}]{shah-etal-2019-robust}
Shah, D.; Gupta, R.; Fayazi, A.; and Hakkani-Tur, D.
\newblock 2019.
\newblock Robust zero-shot cross-domain slot filling with example values.
\newblock In {\em Proceedings of the 57th Annual Meeting of the Association for
  Computational Linguistics},  5484--5490.
\newblock Florence, Italy: Association for Computational Linguistics.

\bibitem[\protect\citeauthoryear{Shi, Fang, and Knight}{2020}]{shi2020dstc}
Shi, X.; Fang, S.; and Knight, K.
\newblock 2020.
\newblock A bert-based unified span detection framework for schema-guided
  dialogue state tracking.
\newblock In {\em Dialog System Technology Challenge Workshop at AAAI}.

\bibitem[\protect\citeauthoryear{Wen \bgroup et al\mbox.\egroup
  }{2017}]{wen2017network}
Wen, T.-H.; Vandyke, D.; Mrk{\v{s}}i{\'c}, N.; Gasic, M.; Barahona, L. M.~R.;
  Su, P.-H.; Ultes, S.; and Young, S.
\newblock 2017.
\newblock A network-based end-to-end trainable task-oriented dialogue system.
\newblock In {\em Proceedings of the 15th Conference of the European Chapter of
  the Association for Computational Linguistics: Volume 1, Long Papers},
  438--449.

\bibitem[\protect\citeauthoryear{Wu \bgroup et al\mbox.\egroup
  }{2019}]{wu-etal-2019-transferable}
Wu, C.-S.; Madotto, A.; Hosseini-Asl, E.; Xiong, C.; Socher, R.; and Fung, P.
\newblock 2019.
\newblock Transferable multi-domain state generator for task-oriented dialogue
  systems.
\newblock In {\em Proceedings of the 57th Annual Meeting of the Association for
  Computational Linguistics},  808--819.
\newblock Florence, Italy: Association for Computational Linguistics.

\bibitem[\protect\citeauthoryear{Xia \bgroup et al\mbox.\egroup
  }{2018}]{xia2018zero}
Xia, C.; Zhang, C.; Yan, X.; Chang, Y.; and Yu, P.
\newblock 2018.
\newblock Zero-shot user intent detection via capsule neural networks.
\newblock In {\em Proceedings of the 2018 Conference on Empirical Methods in
  Natural Language Processing},  3090--3099.
\newblock Association for Computational Linguistics.

\bibitem[\protect\citeauthoryear{Yang \bgroup et al\mbox.\egroup
  }{2019}]{yang2019xlnet}
Yang, Z.; Dai, Z.; Yang, Y.; Carbonell, J.; Salakhutdinov, R.; and Le, Q.~V.
\newblock 2019.
\newblock Xlnet: Generalized autoregressive pretraining for language
  understanding.
\newblock {\em arXiv preprint arXiv:1906.08237}.

\bibitem[\protect\citeauthoryear{Yang, Salakhutdinov, and
  Cohen}{2017}]{yang2017transfer}
Yang, Z.; Salakhutdinov, R.; and Cohen, W.~W.
\newblock 2017.
\newblock Transfer learning for sequence tagging with hierarchical recurrent
  networks.
\newblock {\em arXiv preprint arXiv:1703.06345}.

\bibitem[\protect\citeauthoryear{Zheng, Salvi, and Chan}{2020}]{zheng2020dstc}
Zheng, J.; Salvi, O.; and Chan, J.
\newblock 2020.
\newblock Candidate attended dialogue state tracking using bert.
\newblock In {\em Dialog System Technology Challenge Workshop at AAAI}.

\bibitem[\protect\citeauthoryear{Zhong, Xiong, and
  Socher}{2018}]{zhong-etal-2018-global}
Zhong, V.; Xiong, C.; and Socher, R.
\newblock 2018.
\newblock Global-locally self-attentive encoder for dialogue state tracking.
\newblock In {\em Proceedings of the 56th Annual Meeting of the Association for
  Computational Linguistics (Volume 1: Long Papers)},  1458--1467.
\newblock Melbourne, Australia: Association for Computational Linguistics.

\bibitem[\protect\citeauthoryear{Zhu and Yu}{2018}]{zhu2018concept}
Zhu, S., and Yu, K.
\newblock 2018.
\newblock Concept transfer learning for adaptive language understanding.
\newblock In {\em Proceedings of the 19th Annual SIGdial Meeting on Discourse
  and Dialogue},  391--399.

\end{thebibliography}
\bibliographystyle{aaai}

\end{document}